\documentclass[runningheads]{llncs}
\usepackage[T1]{fontenc}
\usepackage{graphicx}
\usepackage{subcaption}
\begin{document}
\title{Applying Text Embedding Models for Efficient Analysis in Labeled Property Graphs}

\titlerunning{Text Embedding Models for LPGs Analysis}
%
\author{
Michal Podstawski\orcidID{0000-0003-1222-6894}
}

\authorrunning{M. Podstawski}

\institute{
NASK National Research Institute, Warsaw, Poland\\
\url{https://nask.pl/}
\\
\email{michal.podstawski@nask.pl}
}
\maketitle              
\begin{abstract}
Labeled property graphs often contain rich textual attributes that can enhance analytical tasks when properly leveraged. This work explores the use of pretrained text embedding models to enable efficient semantic analysis in such graphs. By embedding textual node and edge properties, we support downstream tasks including node classification and relation prediction with improved contextual understanding. Our approach integrates language model embeddings into the graph pipeline without altering its structure, demonstrating that textual semantics can significantly enhance the accuracy and interpretability of property graph analysis.

\end{abstract}

\section{Introduction}


Graph analysis has become a central tool for modeling and reasoning over complex, interconnected data. In many practical settings, such graphs are enriched with substantial textual information attached to nodes and edges, capturing descriptions, metadata, and contextual details. Labeled property graphs (LPGs) provide a widely used representation for such data, particularly in domains such as knowledge graphs, social networks, and institutional data. Nevertheless, most analytical approaches primarily focus on structural aspects - such as connectivity patterns or edge types - while the semantic content of textual properties remains largely underutilized.

Meanwhile, recent developments in natural language processing have produced powerful pretrained text embedding models capable of capturing nuanced semantic meaning from unstructured text. These models, when applied to LPGs, offer a simple yet effective means of enriching analysis with language-level understanding. Text embeddings can serve as dense, general-purpose representations of nodes and relationships, without requiring any modifications to the underlying graph structure.


We present a lightweight and general framework for integrating pretrained text embeddings into labeled property graph analysis, supporting semantic reasoning without modifying graph topology or retraining models. The approach is evaluated on two core tasks - node classification and link prediction - using semantic vector representations of textual properties as features for simple classifiers. This demonstrates that language-informed embeddings can enhance traditional graph analysis pipelines.

The results suggest that integrating textual semantics into LPG workflows is a practical and scalable strategy for improving the accuracy and interpretability of graph-based analysis.

\section{Related work}


\paragraph{Graph representation learning.}
A large body of work studies how to learn predictive models over graph-structured data.
Early approaches relied on shallow structural embeddings such as DeepWalk and node2vec, which map nodes to vectors by optimizing objectives derived from random walks \cite{perozzi2014deepwalk,grover2016node2vec}.
More recent graph neural networks (GNNs) generalize message passing across neighborhoods and have become standard approach.
Representative architectures include Graph Convolutional Networks (GCN) \cite{kipf2017gcn}, GraphSAGE for inductive learning \cite{hamilton2017graphsage}, Graph Attention Networks (GAT) \cite{velickovic2018gat}, and relational variants such as R-GCN designed for multi-relational graphs \cite{schlichtkrull2018rgcn}.
While these methods typically excel at exploiting topology and typed edges, many practical graphs - including labeled property graphs - contain extensive attribute information (often textual) that is not always leveraged effectively by purely structural encoders.

\paragraph{Attributed and text-rich graphs.}
A parallel line of work addresses graphs with node/edge attributes, including text.
Common strategies concatenate handcrafted text features (e.g., TF-IDF) with structural features, or encode text with neural models and fuse it with GNN representations via concatenation, gating, or attention \cite{yang2016planetoid,zhang2020heterogeneoussurvey}.
There is also work on heterogeneous graphs (with multiple node/edge types) using specialized aggregation and type-aware transformations \cite{wang2019han,hu2019heterogeneous}.
However, these approaches often assume a relatively fixed schema, require end-to-end retraining, or rely on carefully curated feature engineering.
In contrast, labeled property graphs are frequently \emph{schema-flexible}: nodes and relationships can carry varying sets of properties, with missing fields and a mixture of short categorical fields and long free-form text, making uniform feature design and model training more challenging.

\paragraph{Language models for knowledge graphs and graph--text integration.}
Several methods integrate pretrained language models with structured triples for knowledge-graph completion or link prediction.
KG-BERT linearizes a triple (or a small neighborhood) and applies a Transformer to score plausibility \cite{yao2019kgbert}.
KEPLER and related approaches combine language modeling with knowledge embedding objectives, typically requiring joint training and access to curated textual descriptions aligned with knowledge graph entities \cite{wang2021kepler}.
Other lines of work explore text-guided graph encoders, including Graph-BERT-style pretraining on subgraphs and textual signals, or hybrid architectures that encode both local topology and text \cite{zhang2020graphbert}.
Although effective in certain settings, these approaches are usually developed for RDF-like knowledge graphs with fixed predicates, or they impose a particular serialization/training pipeline.
For native property graphs in operational graph databases, a lightweight method that can exploit available text \emph{without} re-engineering the graph schema or training a large joint model remains practically valuable.

\paragraph{Positioning of our approach.}
Compared to end-to-end graph--text fusion methods and KG-specific LM models, we adopt a deliberately lightweight strategy: we serialize available textual properties and local context, encode them with a pretrained text embedding model, and apply standard classifiers for node classification and relation prediction.
This provides a simple interface between language semantics and LPG analytics: it requires no modification of the graph structure, no schema coupling, and no expensive retraining of graph encoders, while still enabling semantic-aware predictions when textual properties carry substantial signal.

\section{Preliminaries}

A labeled property graph (LPG) is a graph data model in which both vertices and edges are first-class entities enriched with semantic annotations~\cite{besta2023demystifying}. Each node and relationship may carry one or more labels that define its conceptual type, as well as an arbitrary set of key--value properties used to store structured or unstructured attributes. Formally, an LPG can be defined as a directed multigraph $G = (V, E, L, P)$, where $V$ is the set of nodes, $E \subseteq V \times V$ is the set of edges, $L$ assigns one or more labels to nodes and edges, and $P$ maps each graph element to a set of properties. This flexible schema allows different nodes and relationships to exhibit heterogeneous structures and attribute sets. Labeled property graphs are widely adopted in practice due to their expressiveness and intuitive alignment with real-world entities and relations, making them well suited for domains such as knowledge graphs, social networks, and institutional data. However, this same flexibility introduces significant analytical challenges: the lack of a fixed schema, the heterogeneity of node and edge types, and the prevalence of high-dimensional and unstructured textual properties make it difficult for traditional graph analytics and learning methods to effectively capture semantic information.

\section{Datasets}

We use publicly available labeled property graph datasets provided by Neo4j~\cite{neo4j}, a widely adopted graph database platform. Used datasets represent diverse domains and graph structures:

\begin{itemize}
    \item Twitter Trolls (nodes 281136, edges 493160)~\cite{neo4j_trolls}: A social network graph capturing interactions between accounts linked to coordinated disinformation campaigns.
    \item Legis (nodes 11825, edges 523004)~\cite{neo4j_legis}: A knowledge graph representation of the U.S. Congress comprising legislators, bills, committees, votes, and related entities.
    \item WWC 2019 (nodes 2486, edges 14799)~\cite{neo4j_wwc2019}: A sports-focused graph dataset modeling the 2019 FIFA Women's World Cup, including players, teams, matches, and events.
    \item Stack Overflow (nodes 6193, edges 11540)~\cite{neo4j_stackoverflow}: A graph modeling Stack Overflow questions, answers, tags, comments, and the relationships between them.
\end{itemize}

These datasets offer rich textual properties on nodes and relationships, making them well-suited for evaluating the integration of text embedding models into graph analysis tasks.

\section{Solution}

Our approach integrates pretrained text embedding models into the analysis of labeled property graphs, enabling semantic-aware workflows for node- and relation-level prediction tasks. The framework is deliberately designed to be model-agnostic: any pretrained text embedding model that maps free-form text to fixed-length vector representations can be substituted without changes to the surrounding pipeline. In this study, we employ Qwen3-Embedding-0.6B~\cite{qwen3embedding} as a representative example, which achieves state-of-the-art performance on the Massive Text Embedding Benchmark~\cite{muennighoff2023mteb,mteb_leaderboard}. This choice enables efficient encoding of textual node and edge properties into dense vector representations without fine-tuning or modifying the underlying graph structure.

At a high level, the proposed pipeline treats textual properties as the primary carrier of semantic information within the graph. For each prediction instance, node-centric information is serialized into a structured textual representation, which is then encoded by the embedding model into a 1024-dimensional vector. These embeddings serve as general-purpose semantic features and are used as input to standard downstream classifiers. This design separates semantic representation learning from task-specific modeling, allowing the same embeddings to be reused across multiple tasks and classifiers.

We consider two prediction tasks, both formulated over node embeddings. In both cases, instances are randomly split into training and test sets, with 90\% of instances used for training and 10\% reserved for evaluation. Downstream prediction is performed using a set of standard classifiers - Random Forest, Logistic Regression, SGDClassifier, and Support Vector Machine (SVM) - trained on the embedding vectors. This choice emphasizes the expressive power of the embeddings themselves rather than task-specific neural architectures, and facilitates transparent comparison across models.

In the \textbf{node label prediction task}, the objective is to predict the semantic label of a node based solely on its textual properties. The input text is constructed by concatenating all available textual properties of the node (excluding the label) into a single description, represented as a sequence of key--value pairs. This formulation reflects a common scenario in property graph analytics, where node types may be unknown or incomplete and must be inferred from descriptive attributes. The classifier learns a mapping from the embedding space to the label space and can subsequently be applied to unseen nodes without requiring access to graph structure or neighborhood information.

\begin{figure}[h!]
\centering
\includegraphics[width=0.7\linewidth]{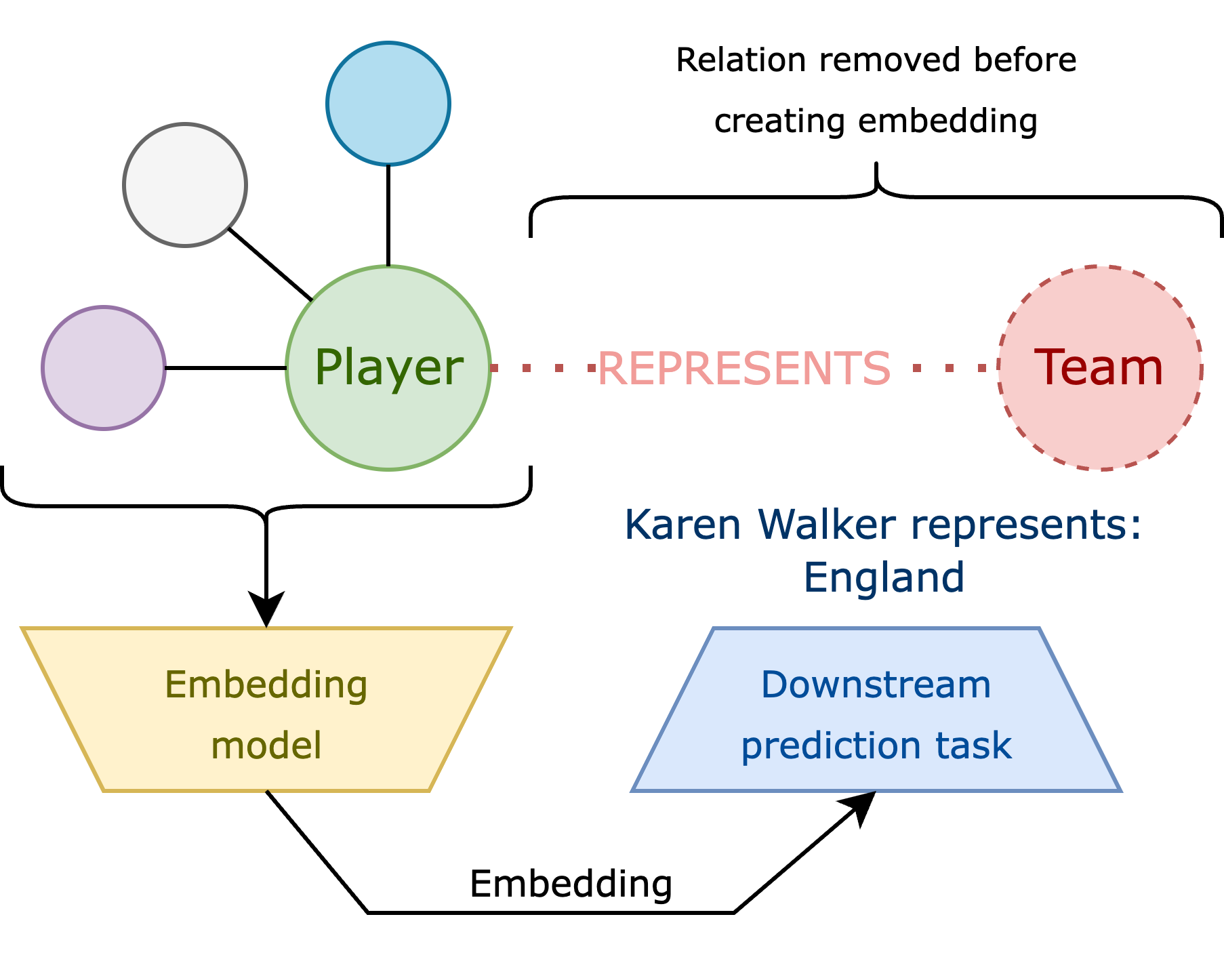}
\caption{\textbf{Relation Prediction Task Setting:} In this setting, a specific relation of the source node (e.g., between a \emph{Player} and a \emph{Team} in the WWC 2019 dataset) is withheld prior to embedding generation. The embedding model encodes the source node based on its remaining relations and neighboring nodes. The resulting embedding is then used in a downstream classification task to predict the correct target node, effectively recovering the withheld relation.}
\label{Fig:task2}
\end{figure}

In the \textbf{relation prediction task}, the goal is to recover missing relational information by leveraging the textual context of a node and its local neighborhood (Figure~\ref{Fig:task2}). For each source node, one relation is withheld prior to embedding generation. The input text is then constructed from the source node’s textual properties, combined with the labels and properties of its remaining relations and directly connected neighbor nodes. This serialized representation captures both intrinsic node descriptions and immediate relational context in textual form. The classifier is trained to predict the correct target node associated with the withheld relation, effectively performing relation recovery based on semantic cues rather than explicit graph traversal.

All textual inputs are normalized to ensure consistency and kept within the token limits of the model. Embeddings are precomputed and cached to enable efficient training and inference. This method requires no modification of the original property graph and remains model-agnostic, allowing substitution with larger or domain-specific embedding models if needed. Overall, the approach allows us to leverage the semantic richness of textual properties in labeled property graphs, enabling more expressive and accurate analysis through modern language model embeddings.


\section{Results}

The proposed approach was applied to property graphs containing varied node types and rich textual attributes. In the node classification task, generated embeddings enabled accurate prediction of labels. Results are presented in Table \ref{Tab:LabelPredictionAll}. For relation prediction, the method successfully recovered information carried by removed edges, based solely on node descriptions and local context. Representative results are gathered in Table \ref{Tab:RelationPredictionAll}. 

The results were consistently strong across datasets, with high classification accuracy and clearly meaningful relation predictions. The model handled both sparse and dense text fields effectively and showed robustness to inconsistencies or missing attributes. Overall, the combination of text embeddings and labeled property graph structure proved to be an effective and efficient foundation for semantic graph analysis.

\begin{table}[h!]
  \centering
  \caption{Classifier performance for predicting node labels across datasets.}
  \label{Tab:LabelPredictionAll}
  \vspace{1em}

    \begin{subtable}{\linewidth}
    \centering
    \begin{tabular}{|l|c|c|c|c|}
      \hline
      Classifier & Accuracy & Precision & Recall & F1 Score \\
      \hline
      Random Forest           & 0.822 & 0.823 & 0.822 & 0.821 \\
      Logistic Regression     & 0.926 & 0.926 & 0.926 & 0.926 \\
      SGDClassifier           & 0.928 & 0.930 & 0.928 & 0.929 \\
      Support Vector Machine  & 0.939 & 0.939 & 0.939 & 0.939 \\
      \hline
    \end{tabular}
    \caption{Stack Overflow}
    \label{Tab:LabelPredictionStackoverflow}
  \end{subtable}

  \vspace{0.6em}
  
  \begin{subtable}{\linewidth}
    \centering
    \begin{tabular}{|l|c|c|c|c|}
      \hline
      Classifier & Accuracy & Precision & Recall & F1 Score \\
      \hline
      Random Forest         & 0.992 & 0.992 & 0.992 & 0.990 \\
      Logistic Regression   & 0.993 & 0.993 & 0.993 & 0.991 \\
      SGDClassifier         & 0.995 & 0.995 & 0.995 & 0.994 \\
      Support Vector Machine & 0.999 & 0.998 & 0.998 & 0.998 \\
      \hline
    \end{tabular}
    \caption{WWC 2019}
    \label{Tab:LabelPredictionWWC}
  \end{subtable}

  \vspace{0.6em}

  \begin{subtable}{\linewidth}
    \centering
    \begin{tabular}{|l|c|c|c|c|}
      \hline
      Classifier & Accuracy & Precision & Recall & F1 Score \\
      \hline
      Random Forest           & 0.984 & 0.984 & 0.984 & 0.983 \\
      Logistic Regression     & 0.999 & 0.999 & 0.999 & 0.999 \\
      SGDClassifier           & 0.994 & 0.993 & 0.994 & 0.994 \\
      Support Vector Machine  & 0.999 & 0.999 & 0.999 & 0.999 \\
      \hline
    \end{tabular}
    \caption{Twitter Trolls}
    \label{Tab:LabelPredictionTrolls}
  \end{subtable}

\end{table}



\begin{table}[!ht]
  \centering
  \caption{Classifier performance for link prediction across datasets.}
  \label{Tab:RelationPredictionAll}
  \vspace{1em}
  
  \begin{subtable}{\linewidth}
    \centering
    \begin{tabular}{|l|c|c|c|c|}
      \hline
      Classifier & Accuracy & Precision & Recall & F1 Score \\
      \hline
      Random Forest         & 0.948 & 0.954 & 0.948 & 0.945 \\
      Logistic Regression   & 0.883 & 0.856 & 0.883 & 0.847 \\
      SGDClassifier         & 0.979 & 0.982 & 0.979 & 0.978 \\
      Support Vector Machine & 0.998 & 0.998 & 0.998 & 0.998 \\
      \hline
    \end{tabular}
    \caption{WCC 2019 - \texttt{REPRESENTS} relation}
    \label{Tab:RelationPredictionWWC}
  \end{subtable}

  \vspace{0.6em}

  \begin{subtable}{\linewidth}
    \centering
    \begin{tabular}{|l|c|c|c|c|}
      \hline
      Classifier & Accuracy & Precision & Recall & F1 Score \\
      \hline
      Random Forest           & 0.796 & 0.807 & 0.796 & 0.786 \\
      Logistic Regression     & 0.858 & 0.875 & 0.858 & 0.851 \\
      SGDClassifier           & 0.919 & 0.914 & 0.919 & 0.916 \\
      Support Vector Machine  & 0.845 & 0.862 & 0.845 & 0.837 \\
      \hline
    \end{tabular}
    \caption{Legis - \texttt{IS\_MEMBER\_OF} relation}
    \label{Tab:RelationPredictionTrolls}
  \end{subtable}

  \vspace{0.6em}

  \begin{subtable}{\linewidth}
    \centering
    \begin{tabular}{|l|c|c|c|c|}
      \hline
      Classifier & Accuracy & Precision & Recall & F1 Score \\
      \hline
      Random Forest           & 0.987 & 0.987 & 0.987 & 0.987 \\
      Logistic Regression     & 0.993 & 0.993 & 0.993 & 0.993 \\
      SGDClassifier           & 0.987 & 0.987 & 0.987 & 0.987 \\
      Support Vector Machine  & 0.987 & 0.987 & 0.987 & 0.987 \\
      \hline
    \end{tabular}
    \caption{Legis - \texttt{ELECTED\_TO} relation}
    \label{Tab:RelationPredictionStackoverflow}
  \end{subtable}

\end{table}

\section{Limitations and Scope}

While the proposed approach demonstrates strong performance across multiple datasets and tasks, it is important to clarify its scope and limitations. First, the effectiveness of the method depends on the availability and quality of textual properties associated with nodes and relationships. In graphs where textual attributes are sparse, highly noisy, or semantically uninformative, the benefit of text embeddings may be limited. In such cases, structural signals or task-specific features may play a more dominant role.

Second, the approach intentionally does not incorporate explicit structural encoders or message-passing mechanisms. While this design choice enables a lightweight and model-agnostic integration into existing labeled property graph pipelines, it also means that higher-order structural patterns are not directly captured by the embedding process. Tasks that rely heavily on global topology or multi-hop relational reasoning may therefore require complementary structural models or hybrid approaches that combine language-based embeddings with graph-aware encoders.

Finally, the current evaluation focuses on node classification and relation prediction as representative analysis tasks. While these tasks cover common use cases in property graph analytics, further evaluation on additional tasks - such as clustering, anomaly detection, or semantic search - would provide a more comprehensive assessment of the approach’s generality. Despite these limitations, the proposed framework offers a practical and extensible foundation for incorporating textual semantics into labeled property graph analysis without modifying existing graph structures or workflows.

\section{Next Steps}

Several directions emerge for extending the proposed framework. A natural next step is the integration of textual embeddings with topology-aware graph encoders. While the current approach deliberately separates semantic representation from graph structure, combining language-model embeddings with graph neural networks or other structural encoders could enable joint reasoning over textual content and multi-hop relational patterns. Such hybrid models may be particularly beneficial for tasks that depend on global graph structure or complex relational dependencies.

Another promising direction concerns task-specific adaptation of the embedding space. Although pretrained text embedding models provide strong general-purpose representations, lightweight fine-tuning or adaptation strategies could further improve performance in domain-specific settings. This includes adapting embeddings to specialized vocabularies, domain terminology, or task objectives commonly encountered in institutional or knowledge-centric graphs.

Scalability and dynamic graph support also represent important avenues for future work. Large labeled property graphs are often subject to frequent updates, including the addition of new nodes, relationships, and textual properties. Developing efficient mechanisms for incremental embedding updates and embedding lifecycle management would be essential for supporting real-time or near-real-time analytics.

\section{Conclusions}

We presented an approach for enhancing labeled property graph analysis using pretrained text embedding models. By encoding textual properties into semantic vectors, we enable more informed classification and relation prediction without modifying the underlying graph structure. This method is model-agnostic, lightweight, and compatible with existing property graph platforms. Our evaluation demonstrates its broad applicability across domains.

The results highlight the potential of treating textual node and edge attributes as semantically rich signals, supporting more expressive and effective analysis workflows within graph-based systems. This framework shows that semantic text embeddings can serve as drop-in, model-independent enhancements for graph analytics-bridging language understanding and structural reasoning in a unified, efficient manner.

\begin{credits}
\subsubsection{\ackname} This manuscript acknowledges the use of ChatGPT~\cite{chatgpt}, powered by the GPT-5.2 language model developed by OpenAI, to improve language clarity, refine sentence structure, and enhance overall writing precision.
\end{credits}
%
%
%
%

\end{document}